\documentclass[journal]{IEEEtran}

\usepackage{xcolor,soul,framed} 

\colorlet{shadecolor}{white}
\usepackage[pdftex]{graphicx}
\graphicspath{{../pdf/}{../jpeg/}}
\DeclareGraphicsExtensions{.pdf,.jpeg,.png}

\usepackage[cmex10]{amsmath}
\usepackage{array}
\usepackage{mdwmath}
\usepackage{mdwtab}
\usepackage{eqparbox}
\usepackage{url}

\usepackage{amssymb,amsfonts}
\usepackage{graphicx}
\usepackage{algorithm}
\usepackage{hyperref}
\usepackage{cite}
\hypersetup{hidelinks=true}
\usepackage{textcomp}
\usepackage{makecell}
\usepackage{multirow}
\usepackage{float}
\usepackage{algpseudocode}
\usepackage{array}

\begin{document}
\title{MMeViT: Multi-Modal ensemble ViT for Post-Stroke Rehabilitation Action Recognition}
\author{Ye-eun Kim, Suhyeon Lim, and Andrew J. Choi
\thanks{
This work was supported by the Translational Research Center 
for Rehabilitation Robots(NRCTR-EX23008), National Rehabilitation Center, Ministry of Health and Welfare, Korea and the Gachon University research fund of 2024(GCU-202400480001). 
(Ye-eun Kim and Suhyeon Lim contributed equally to this work.) (Corresponding author: Andrew J. Choi)
} 

}

\markboth{arXiv preprint
}{Y. Kim \MakeLowercase{\textit{et al.}}: MMeViT: Multi-Modal ensemble ViT for Post-Stroke Rehabilitation Action Recognition}

\maketitle

\begin{abstract}
Rehabilitation therapy for stroke patients faces a supply shortage despite the increasing demand. 
To address this issue, remote monitoring systems that reduce the burden on medical staff are emerging as a viable alternative. A key component of these remote monitoring systems is Human Action Recognition (HAR) technology, which classifies actions. However, existing HAR studies have primarily focused on non-disable individuals, making them unsuitable for recognizing the actions of stroke patients. HAR research for stroke has largely concentrated on classifying relatively simple actions using machine learning rather than deep learning. In this study, we designed a system to monitor the actions of stroke patients, focusing on domiciliary upper limb Activities of Daily Living (ADL). Our system utilizes IMU (Inertial Measurement Unit) sensors and an RGB-D camera, which are the most common modalities in HAR. We directly collected a dataset through this system, investigated an appropriate preprocess and proposed a deep learning model suitable for processing multimodal data. We analyzed the collected dataset and found that the action data of stroke patients is less clustering than that of non-disabled individuals. Simultaneously, we found that the proposed model learns similar tendencies for each label in data with features that are difficult to clustering. This study suggests the possibility of expanding the deep learning model, which has learned the action features of stroke patients, to not only simple action recognition but also feedback such as assessment contributing to domiciliary rehabilitation in future research. The code presented in this study is available at \underline{https://github.com/ye-Kim/MMeViT}.
\end{abstract}


\section{Introduction}

\IEEEPARstart{P}{ost}-stroke patients with hemiparesis often exhibit a tendency to rely predominantly on their unaffected, non-paretic limb. The recovery of motor function in the paretic limb is contingent upon its consistent and sustained use. In addition to supervised clinical sessions, it is crucial to encourage patients to spontaneously use their paretic limb in daily life to accelerate functional recovery \cite{gomez-arrunategui_monitoring_2022}. Consequently, there has been a growing interest in developing action monitoring systems for patients to recognize movements of patients and provide real-time feedback and personalized therapy. Due to the increasing demand for rehabilitation training in daily life to enhance its effectiveness and the need to alleviate the burden on healthcare professionals due to the supply shortage, the demand for monitoring systems that support autonomous exercise for stroke patients is also rising \cite{gomez-arrunategui_monitoring_2022}, \cite{collins_stroke_2017}. In such rehabilitation training monitoring, human action recognition for patients is a critical component \cite{shen_rarn_2022}.

Human action recognition (HAR) is a field that automatically identifies, classifies, and interprets human actions using vision data collected by camera \cite{kadambi_detecting_2025}, \cite{filtjens_automated_2022}, or signal data collected by sensors \cite{zubair_human_2016}, \cite{liu_human_2024}. Human action recognition is utilized and researched in various fields. In particular, multimodal HAR has been studied to improve the performance of HAR by using multiple modalities to compensate for the shortcomings that may arise from using a single modality \cite{ahn_multimodal_2024}, \cite{sun_human_2022}. However, most existing multimodal HAR studies have focused on the movements of non-disabled individuals. In the field of stroke patient movement recognition, due to the difficulty of having patients wear numerous sensors, cameras have been predominantly used \cite{hardPostStroke}. Furthermore, research has primarily focused on classifying lower limb, especially gait \cite{filtjens_automated_2022}, \cite{guo_physics-informed_2025}, \cite{xu_graph-based_2025}, rather than upper limbs, and there has been a tendency to use simple unimodal data and machine learning rather than deep learning.

This study established the foundation for a real-time movement monitoring system by building a post-stroke indoor upper limb-focused Activities of Daily Living (ADL) monitoring system and performing HAR with multimodal data directly collected through this system. To minimize user discomfort, the number of sensors and cameras worn by the participants was minimized. The action of stroke patients are less clustering compared to those of non-disabled individuals. Nevertheless, we developed an initial deep learning model that learns the tendencies of movement data performed by stroke patients. We also proposed an appropriate preprocess that quickly converts sensor data into images 
and extracts skeletons from the RGB camera, which is suitable for multimodal stroke HAR. This study proposes the potential of multimodal HAR deep learning for post-stroke subjects by enabling the ensemble of this preprocessed multimodal data using the same model.

This research is significant because it directly collected upper limb ADL data from stroke patients and performed movement recognition. The biggest obstacle in HAR systems for stroke patients is the difference in movement characteristics between the subjects of existing research and the actual HAR users. Human upper limb action is complex. Compared to lower limb actions like walking or standing up, which have a simple set of movements, upper limb actions for stroke patients have more complex movements with a wider variety of ADL and Range of Motion (ROM) actions. Therefore, for effective HAR, the position of the sensors and cameras is as important as the type of modality used \cite{hardPostStroke}.

To maximize the capture of upper limb movement characteristics, we had the participants wear one IMU sensor on each wrist. We collected movement data from stroke patients and performed HAR by linking the data with a camera placed in front of the participants to supplement the skeleton information that is difficult to obtain with only the IMU sensors. In this process, we attempted to improve accuracy using an ensemble ViT model \cite{ViT}, and it is expected that using skeleton data will also enable to contribute to domiciliary rehabilitation in future research.

\section{Relative Works}

\subsection{Action Recognition Based on RGB(D) Data}
Using a single or multiple of RGB-D camera for HAR provides rich visual information and simplifies the process of collecting and preprocessing data due to the availability of various developed libraries. RGB(D) data is usually easy to collect ,and it contains rich appearance information of the captured scene context. Most of the existing works focused on using videos to handle HAR \cite{kadambi_detecting_2025}, \cite{proffitt_development_2023},\cite{kamal_novel_2025}. There are few studies that have been conducted in frame units \cite{zhang_frame_2025}.

However, it has the drawbacks of potential occlusion depending on the shooting angle. HAR from RGB data is often challenging because of the necessity of a preprocessing step that of extracting only the human part from the data owing to the varitaitons of backgrounds, viewpoints, and illumination conditions.

\subsection{Action Recognition Based IMU Sensor}
Sensor-based HAR offers advantages such as providing detailed information, protecting personal privacy, and having a relatively smaller data size compared to visual modalities. However, it lacks visual appearance information and can only capture data from the area where the sensor is attached.

In sensor-based HAR, IMU sensors (accelerometer and gyrometer) \cite{gomez-arrunategui_monitoring_2022}, \cite{zubair_human_2016}, \cite{oh_data_2024,jin_deep_2024,smith_detecting_2024,oh_investigating_2023,orlov_features_2019,rahman_continual_2023,oubre_estimating_2020,werner_using_2022}, EMG (electromyography) and sEMG(surface EMG) sensors \cite{lee_decoding_2025,bao_deep_2025,zhang_multilevel_2025,makeSensor} are primarily used. Unlike EMG sensors, which must be attached directly to the skin, IMU sensors can be worn over clothing. This ease of use makes IMU sensors particularly suitable for collecting movement data from stroke patients who may have muscle rigidity.

Traditionally, HAR using IMU sensors extracts a single representative time-series value \cite{oh_data_2024}, \cite{oh_investigating_2023}, \cite{orlov_features_2019}, \cite{minarno_single_2020}, such as the vector magnitude of the 3-axis (x, y, z) data from each modality. However, when using multiple sensors and combining them with other modalities \cite{liu_human_2024}, this approach can lead to a loss of data features that could be correlated with the other modalities.

The Action Recognition discussed above is mostly for non-disabled individuals. Since the movements of stroke patients differ from those of non-disabled individuals \cite{hardPostStroke}, there is a growing need for HAR research tailored to the unique movement characteristics of stroke patients.

\subsection{Action Recognition for Stroke Patients}
The field of post-stroke patient HAR is gaining attention for patient action monitoring \cite{gomez-arrunategui_monitoring_2022}, \cite{collins_stroke_2017}, \cite{moore_depth_2019}, \cite{miao_upper_2021}, \cite{ma_vicovr-based_2018}. Consistent rehabilitation, along with the possibility of remote monitoring and telemedicine, can reduce the burden on medical staff and allow patients to undergo rehabilitation at home rather than in a specialized medical facility. In addition to post-Stroke patients, HAR of Parkinson's disease or brain lesion patients has been conducted \cite{filtjens_automated_2022}, \cite{xu_graph-based_2025}, \cite{smith_detecting_2024}.

However, while existing research has proposed systems for patient HAR, few studies have actually performed HAR using real patient data. Gait classification studies for stroke patients are predominant \cite{guo_physics-informed_2025}. But gait classification typically involves only two labels—walking and stopping—and can be relatively easily classified with just two IMU sensors worn on the ankle or thigh, without even needing a video for multimodal processing.

Action recognition for stroke patients primarily focuses on indoor settings. Systems for ADL recognition in post-stroke rehabilitation have been studied using depth sensors and RGB-D cameras \cite{deb_graph_2022}, \cite{kuang_hierarchical_2025}. There are also studies that use IoT to classify ADL actions \cite{jin_deep_2024}, but because the models are trained on non-disabled individuals' data rather than a stroke patient dataset \cite{kadambi_detecting_2025}, the accuracy of the action recognition on data collected from an actual system is unknown.

\subsection{Multivariate Time Series Data Classification}
Sensor data is a time-series dataset with multiple channels, and skeletons extracted on a frame-by-frame basis also fall under multivariate time-series data. Unlike single time series data classification, the goal of classifying multivariate time series data is to enhance accuracy by considering the relationships between variables \cite{liu_human_2024}, \cite{xu_graph-based_2025}, \cite{jin_deep_2024}, \cite{capela_evaluation_2016}, \cite{kantoch_human_2017}.

Models such as InceptionTime and Conv1D, which are based on CNNs (Convolutional Neural Networks), and ConvLSTM, which combines CNNs with RNN (Recurrent Neural Network)-based models like BiLSTM, have been developed for this purpose \cite{hardPostStroke}.

To maximize the relationships between variables and leverage high-performing image recognition models from the field of deep learning, methods for representing time-series data as images have also been devised. Techniques for visualizing time-series data include Recurrence Plot, Gramian Angular Field, Markov Transition Field, and Spectrogram \cite{ts2img}.

\subsection{Skeleton Heatmap}
Skeletons inferred from RGB-D data \cite{shen_rarn_2022}, \cite{kuang_hierarchical_2025}, \cite{liu_revealing_2025}, \cite{li_study_2015}, have the advantage of being simpler and providing only the necessary information compared to raw video data, but they tend to be noisy \cite{ahn_multimodal_2024}, \cite{guo_physics-informed_2025}, \cite{STGCN}, \cite{duan2022pyskl}. This is especially true when a non-depth camera is used, as noise is often generated during the depth estimation process. In this study, to minimize the impact of this noise on classification, we created a heatmap by estimating the z-axis using a Gaussian filter \cite{deb_graph_2022}.

In action recognition, a skeleton fundamentally represents a person's position within a single frame. However, action recognition requires positional information over time. Therefore, various models have been researched to effectively classify continuous skeleton data. Among existing studies, pyskl \cite{duan2022pyskl} offers comparable performance to large-scale models like LLM (Large Language Model), and its manageable model size makes it a suitable approach for this research, particularly for dissemination to stroke patients.

\section{Methods}
\label{sec:Methods}

\subsection{Collecting Dataset}

\begin{figure}[h]
\centerline{\includegraphics[width=\columnwidth]{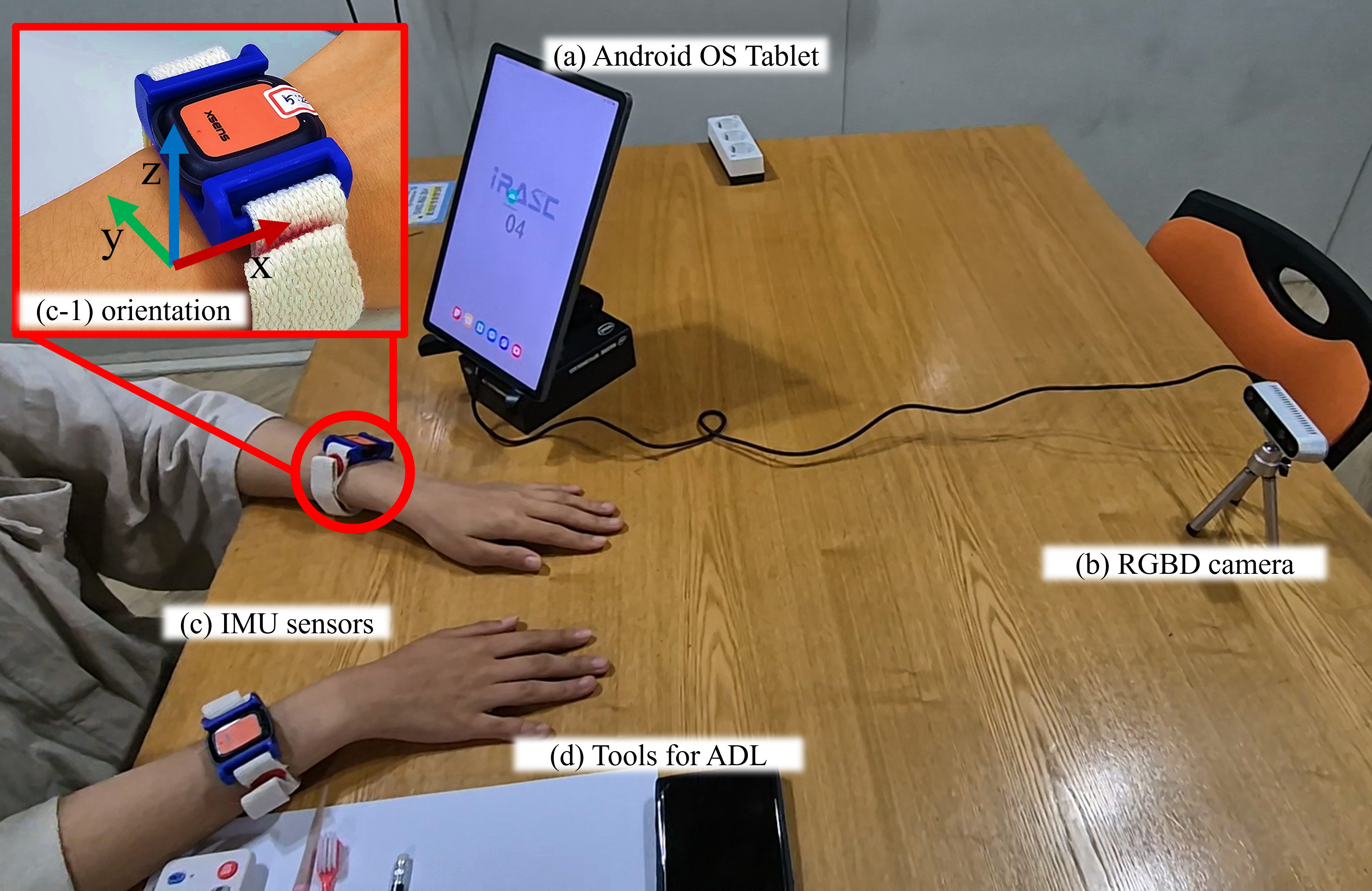}}
\caption{System setting. (a) Android tablet to synchronize the camera and IMU sensor, (b) RGB-D camera, (c) IMU sensors and (c-1) the orientation of IMU sensors,
(d) tools for ADL} 
\label{fig1} 
\end{figure}

The dataset was collected as shown in Figure \ref{fig1}. This system for collecting dataset was designed like activity monitoring system \cite{collins_stroke_2017}, \cite{li_study_2015}. The camera (Figure \ref{fig1}(b)) and sensors (Figure \ref{fig1}(c)) were synchronized via an Android tablet (Figure \ref{fig1}(a)). The sensor orientation was set to Figure \ref{fig1}(c-1), and data was collected simultaneously while ensuring time step alignment. The user then performed the presented actions using tools (Figure \ref{fig1}(d)) at the speed of each participant without a specific time limit. The presented actions were mostly complex ADLs that required using tools. The list of action labels collected from the dataset is as Table \ref{tab1}.

\begin{table}[h]
\caption{Activities of Daily Living (ADL) List}
\resizebox{\columnwidth}{!}{
\begin{tabular}{cp{50pt}p{110pt}p{35pt}}
\Xhline{2\arrayrulewidth}
\textbf{Index} & \textbf{Label} & \textbf{Description} & \textbf{Tool}
\\
\Xhline{2\arrayrulewidth}
1 & LiftCupHandle & Drinking Water in a cup with handle & cup \\
2 & HairBrush & Brushing hair with a hairbrush & hairbrush \\
3 & BrushTeeth & Brushing Teeth with a toothbrush & toothbrush \\
4 & Remotecon & Pressing the remote control button & remote controller \\
5 & MovingCan & Moving a can without handle & can \\
6 & Writing & Writing with one hand with holding the paper with the other hand & pencil, paper \\
7 & FoldingPaper & Folding a paper & paper \\
8 & FoldUpTower & Folding a towel & towel \\
9 & WashFace & Washing face without water & None \\
\Xhline{2\arrayrulewidth}
\end{tabular}
}
\label{tab1} 
\end{table}

The collected data types were IMU (3-axis accelerometer, gyrometer) and RGB video data. We recruited a total of 84 participants (76 healthy, non-disabled individuals(ND) aged 20-31, and 18 stroke hemiplegia patients(Stroke) aged 27-87 with onset times ranging from 1 to 55 months) for the data collection experiment.

\begin{table}[h]
\caption{Stroke Participants} 
\resizebox{\columnwidth}{!}{
\scriptsize 
\begin{tabular}{c|cccccc}
\Xhline{2\arrayrulewidth}
ID & \makecell{Handedness \\ (L/R)} & \makecell{Affected Side \\ (L/R)} 
& \makecell{Gender \\ (F/M)} & \makecell{Age \\ (years)} 
& \makecell{Stroke Onset \\ (Months)} & \makecell{MAS \\ (L/R)} \\
\Xhline{2\arrayrulewidth}
Stroke01 & R & L & F & 70 & 2 & G0/G0 \\
Stroke02 & R & R & F & 75 & 2 & G1/G0 \\ 
Stroke03 & R & R & F & 56 & 3 & G1/G0 \\ 
Stroke04 & R & L & F & 81 & 8 & G1/G0 \\ 
Stroke05 & R & L & M & 56 & 6 & G1/G0 \\ 
Stroke06 & R & R & M & 69 & 3 & G1/G0 \\ 
Stroke07 & R & L & M & 69 & 4 & G1/G0 \\ 
Stroke08 & R & L & F & 77 & 6 & G1/G0 \\ 
Stroke09 & R & L & F & 64 & 2 & G1/G0 \\ 
Stroke10 & R & L & M & 27 & 4 & G1/G0 \\ 
Stroke11 & R & L & F & 89 & 9 & G1/G0 \\ 
Stroke12 & R & L & M & 74 & 8 & G1/G0 \\ 
Stroke13 & R & R & F & 65 & 17 & G1/G0 \\ 
Stroke14 & R & R & M & 45 & 56 & G1/G0 \\ 
Stroke15 & R & R & F & 81 & 56 & G1/G0 \\ 
Stroke16 & R & R & F & 66 & 68 & G1/G0 \\ 
Stroke17 & R & R & M & 83 & 16 & G1/G0 \\ 
Stroke18 & R & R & M & 70 & 9 & G1/G0 \\ 
\Xhline{2\arrayrulewidth}
\end{tabular}
}
\label{tab2} 
\end{table}

All ND participants completed 8 sessions each, while Stroke participants completed 3-18 sessions depending on their individual condition. The pathological information for the Stroke participants is shown in Table \ref{tab2}.

\begin{table}[h]
\caption{Statistics of Segment Lengths}
\resizebox{\columnwidth}{!}{
    \scriptsize 
    \begin{tabular}{ll!{\vrule width 1pt}ccccc}
    \Xhline{2\arrayrulewidth}
            &     & mean & std & min & med & max \\
    \Xhline{2\arrayrulewidth}
    \multirow{2}*{IMU} & ND & 414 & 237 & 151 & 531 & 2386 \\
                 & Stroke & 904 & 649 & 155 & 709 & 4343 \\
    
    \hline 
    {RGB-D} & ND & 411 & 235 & 100 & 344 & 2852 \\
    {  video} & Stroke & 570 & 519 &  90 & 412 & 5464 \\
    \Xhline{2\arrayrulewidth}
    
    \end{tabular}
}
std = standard deviation, med = median value of the segment lengths.
\label{tab3}
\end{table}

After removing corrupted data, a total of 6,293 segments were used for training the deep learning model in this study (4,859 for ND and 1,434 for Stroke). The length distribution of these collected data segments is presented in Table \ref{tab3}. In Table \ref{tab3}, std means standard deviation and med means median.

This experiment for collecting dataset was approved by the IRB of Gachon university before proceeding (IRB no. 1044396-202306-HR-102-02). The experimental design, action selection, and execution for the dataset collection were conducted with the consultation and inspection of three therapists with bachelor's degrees in physical or occupational therapy and over five years of experience, as well as two rehabilitation Ph.Ds.

\begin{figure*}[h]
\centerline{\includegraphics[width=\textwidth]{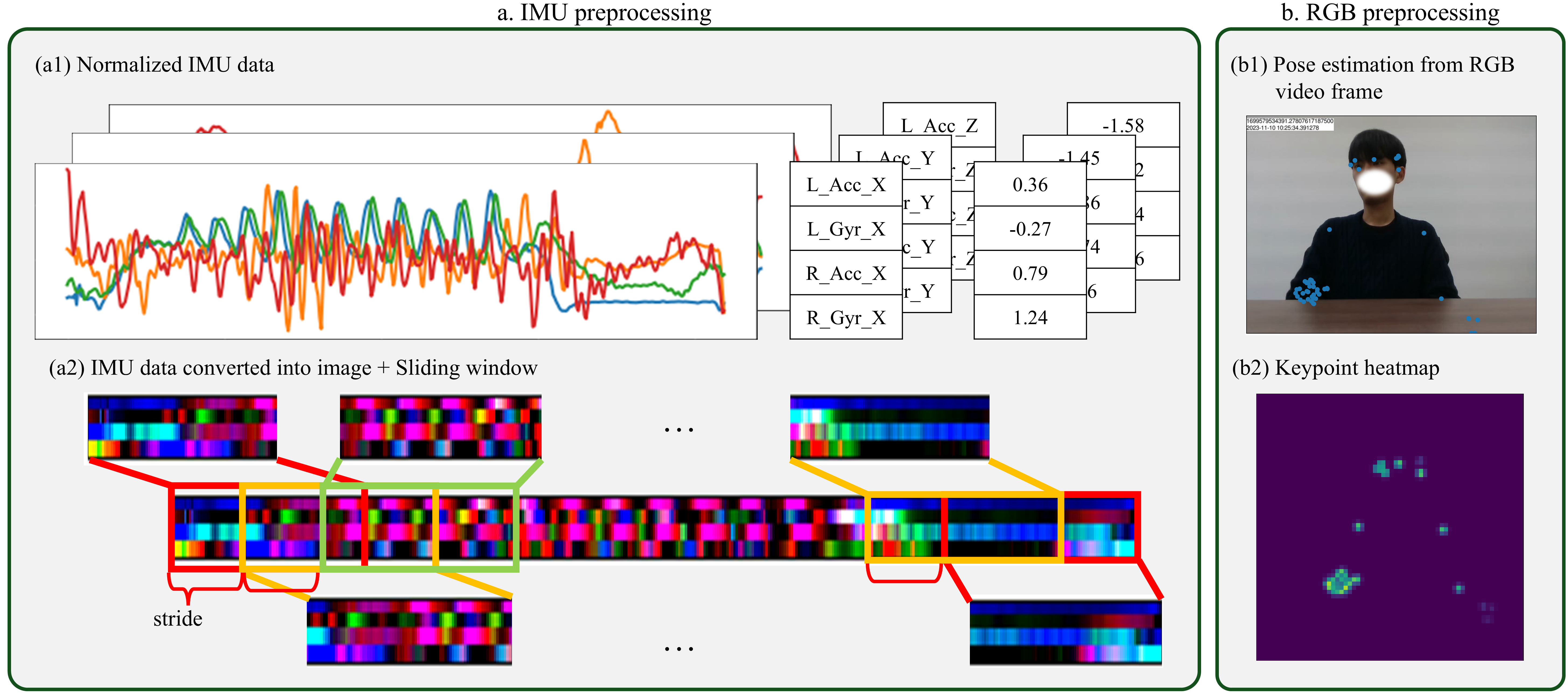}}
\caption{Data Preprocessing Overview. (a) Preprocessing of IMU data is composed (a1) normalization and (a2) converting into image and sliding window. (b) Preprocessing of RGB data is composed (b1) extracting skeleton of person from every frame using pose estimation and (b2) make the skeleton keypoint heatmap.}
\label{fig2}
\end{figure*}

\subsection{Data Preprocessing}

\subsubsection{Image representation of IMU data}
To use ViT model, aceelerometer and gyrometer data collected by IMU sensor were converted into 3-channel 2-dimension image after z-score normalization. Then a sliding window approach was applied to ensure consistency in the input dimensions. The formula of z-score normalization (Figure \ref{fig2} a1) is as follows.

\begin{equation} 
\mathit{Acc}_{ij} = \frac{\mathit{Acc}_{ij} - \mu_{\mathit{acc}}}{\sigma_{\mathit{acc}}}, \label{equa1}
\end{equation}

\begin{equation}
\mathit{Gyr}_{ij} = \frac{\mathit{Gyr}_{ij} - \mu_{\mathit{gyr}}}{\sigma_{\mathit{gyr}}} \label{equa2}
\end{equation} 
where \(i \in \{\mathit{left hand}, \mathit{right hand}\}\) denotes the sensor location 
and \(j \in \{x, y, z\}\) represents the three axis of sensor data.

We reorganized the original matrix of size \([T, 2 \times M \times 3]\) by interpreting each axis as an independent channel, analogous to the RGB channels in an image (Figure \ref{fig2} a2). This results in a matrix of size \([T, 2 \times M, 3]\), where \(T\) denotes the temporal length (number of frames), and \(M\) indicates the number of sensor modalities (accelerometer and gyrometer).

Then, to adjust the input data size, we used sliding window \cite{dentamaro_human_2024}. The number of windows \(n\) is calculated as follows:
\begin{equation}
n = 1 + \left\lfloor \frac{T - W}{S} \right\rfloor \label{equa3}
\end{equation}
where \(n\) is the number of windows, \(T\) denotes the temporal length, \(W\) is the window length, 
and \(S\) is the stride. In this study, the window length \(W\) is set 120.

\subsubsection{Skeleton heatmap from RGB Video}

\begin{figure}[h]
\centerline{\includegraphics[height=4cm]{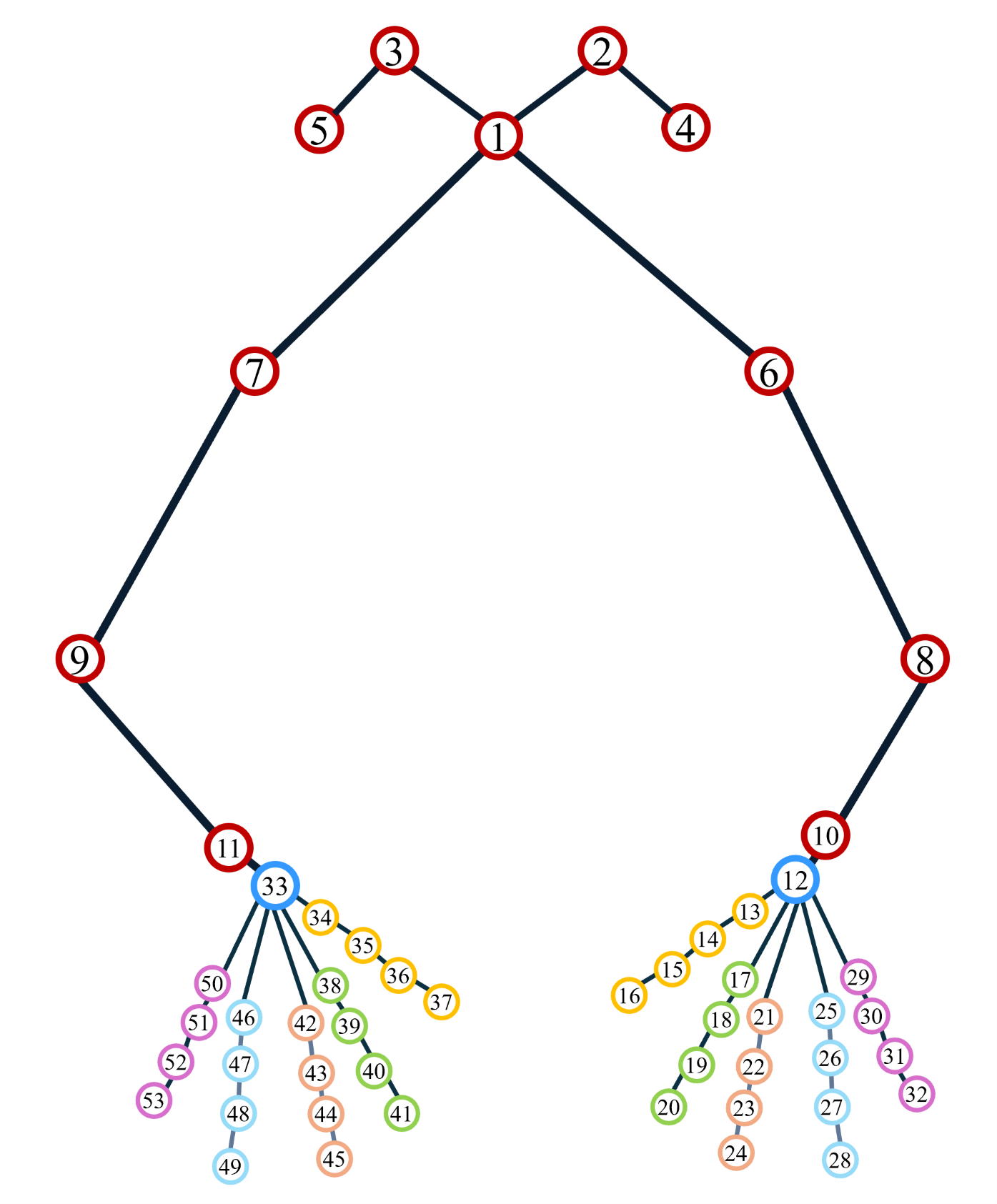}}
\caption{Selected 53 Skeleton Keypoints}
\label{fig3} 
\end{figure}

To perform action recognition, we used the pose estimation library MMAction2 \cite{2020mmaction2} to extract keypoint coordinates from each frame of the collected RGB video. The extracted keypoints consist of 53 main body parts like Figure \ref{fig3}, including both hands and the upper body, which form a heatmap annotation. As the extracted keypoints are 2D coordinates, they were re-constructed into a 3D heatmap \cite{liu_revealing_2025}, \cite{STGCN}. Then we used frame drop techniques to adjust the input data size. 

We construct a gaussian map centered at each of the \(k\) keypoints to obtain the keypoint heatmap \(J\). Formally, the heatmap is defined as:

\begin{equation}
\label{equa4}
J_{kij} = \exp\left(-\frac{(i - x_k)^2 + (j - y_k)^2}{2\sigma}\right) \cdot c_k
\end{equation} 

where \(\sigma\) denotes the variance of the Gaussian, \((x_k, y_k)\) are the coordinates of the \(k\)-th joint, and \(c_k\) represents the confidence score of the corresponding keypoint. This formulation allows the heatmap to emphasize spatial regions around each keypoint, while the confidence score modulates the overall contribution of each joint.

Since each keypoint is defined for every frame, the heatmaps \(J\) are aggregated along the time axis to construct a 3D heatmap of size \([K, T, H, W]\), where \(K\) denotes the number of keypoints, \(T\) is the temporal length (number of frames), \(H\) represents the height, and \(W\) represents the width.

\begin{figure*}[ht!]
\centerline{\includegraphics[width=\textwidth]{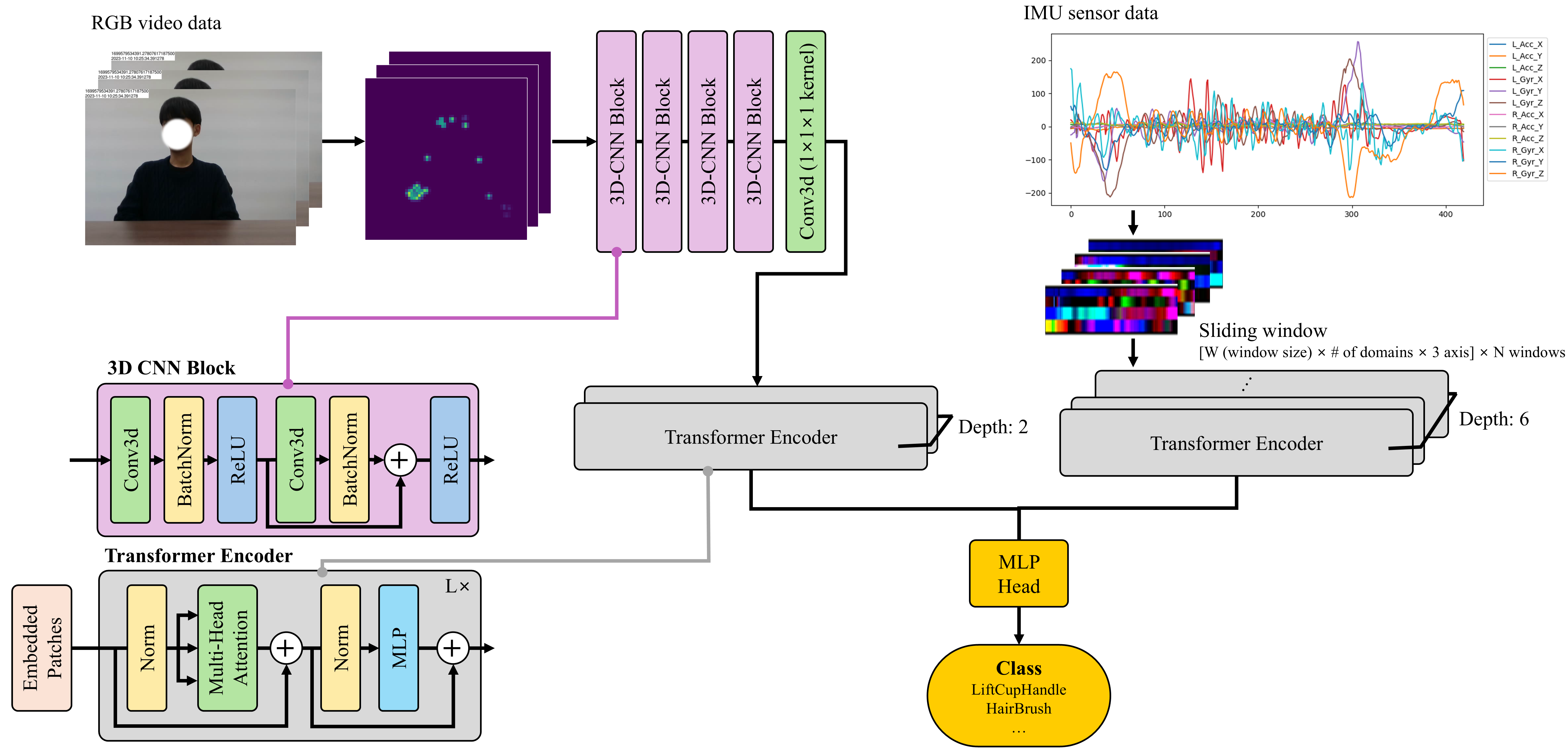}}
\caption{Ensemble Model Overview} 
\label{fig4} 
\end{figure*}

While traditional data augmentation techniques focus on increasing the number of segments \cite{oh_data_2024}, \cite{minarno_single_2020}, a simultaneous augmentation method for the paired IMU-skeleton data used in this study has not yet been researched, as the skeleton data is one dimension higher than the IMU data.

Therefore, during model training, we applied flip and crop on a frame-by-frame basis to the skeleton data. To increase data diversity, we can also shift frames from consecutive sequences (e.g., by selecting only the first, second, or a random frame from sub-segments) or randomly sample them. For the IMU data, we augmented features within the segment using a sliding window approach \cite{dentamaro_human_2024}.

\subsection{Ensemble Model}

The overall ensemble model is like Figure \ref{fig4}.
The heatmap data generated from the video is input into a model composed of 3D CNN blocks. 
The initial four stages of the total 5-stage 3D-CNN block are ResNet blocks, which effectively extract spatiotemporal information. Subsequently, this information is extracted, compressed, and converted into a 1-channel 2D gray-scale feature map.

The ViT (Vision Transformer) architecture \cite{ViT} is strong for image-based classification and enables global learning. This 2D feature map is divided into fixed-size patches and passed through a transformer block with a self-attention mechanism to represent the correlation between each region as a representation vector (depth=2).

The accelerometer and gyroscope data obtained from the IMU sensors are first transformed into 3-channel 2D windows. To capture the inter-domain relationships, each window is further divided into patches of size of \(\mathit{number of domains} \times \mathit{number of domains}\), which are then fed into a transformer block with a self-attention mechanism. This enables the temporal correlations across all domains to be encoded into a representation vector. The transformer consists of 6 layers (depth=6).

The spatiotemporal and time-unit correlations of acceleration and angular velocity are represented 
by their respective representation vectors. These two vectors are then concatenated to compute a single, final representation vector that represents the features of the input action. This final representation vector is fed into a separately trained MLP head for classification.

This multimodal ensemble model is constructed by utilizing the pre-trained ViT model with each unimodal and training a small number of MLP head for a few epochs. The MLP head consists of one layer normalization layer, one FC layer, and one softmax layer, performing the classification using only the representation vector. Cross-entropy loss was used as the criterion.

\section{Experiments}
\label{sec:Experiments}

\subsection{Experiment Settings}

The implementation and training were conducted on an Intel i9-10920X CPU and an NVIDIA GeForce RTX 4090 GPU. All model frameworks were developed using PyTorch on Ubuntu 22.04.05 LTS. 

\begin{table}[ht]
\centering
\caption{The number of Action segments of Collected Dataset} 
\begin{tabular}{c!{\vrule width 1pt}ccc|c}
\Xhline{2\arrayrulewidth}
\textbf{data} & \textbf{train} & \textbf{valid} & \textbf{test} & \textbf{total} \\
\Xhline{2\arrayrulewidth}
ND & 3401 & 973 & 485 & 4859 \\
Stroke & 1003 & 288 & 143 & 1434 \\
\hline 
total & 4404 & 1261 & 628 & 6293 \\
\Xhline{2\arrayrulewidth}
\end{tabular}

\label{tab4} 
\end{table}

The collected dataset was split into training, validation, and test sets in a 7:2:1 ratio, and the number of segments used in each set is summarized in Table \ref{tab4}.

\subsection{Hyperparameter Settings}
The Adam optimzier was used to train all models, and the hyperparameters set in each model learning process are as Table \ref{tab5}.

\begin{table}[ht]
\caption{Hyperparameter settings for each models}
\centering
\resizebox{\columnwidth}{!}{
\begin{tabular}{c!{\vrule width 1pt}c|c|c}
\Xhline{2\arrayrulewidth}
\textbf{Model} & \textbf{ViT-IMU} & \textbf{ViT-Skeleton} & \textbf{MLP} \\
\Xhline{2\arrayrulewidth}
Batch size & [8, 16, \underline{32}, 64] & [4, 8, 16, \underline{32}] & [8, 16, \underline{32}, 64] \\
Learning rate & [\underline{1e-5}, 5e-5, 1e-4, 5e-4] & [\underline{1e-4}, 5e-4, 1e-3, 5e-3] & [5e-5, 1e-4, 5e-4, \underline{1e-3}] \\
Train epoch & [10, \underline{40}, 50, 100] & [10, 20, 50, \underline{100}] & [\underline{10}, 20, 40, 100] \\
\Xhline{2\arrayrulewidth}
\end{tabular}
}
\label{tab5}
\end{table}

\section{Results}
\label{sec:Results}

\subsection{Result of Experiment}

\begin{figure}[!h]
\centerline{\includegraphics[width=0.8\columnwidth]{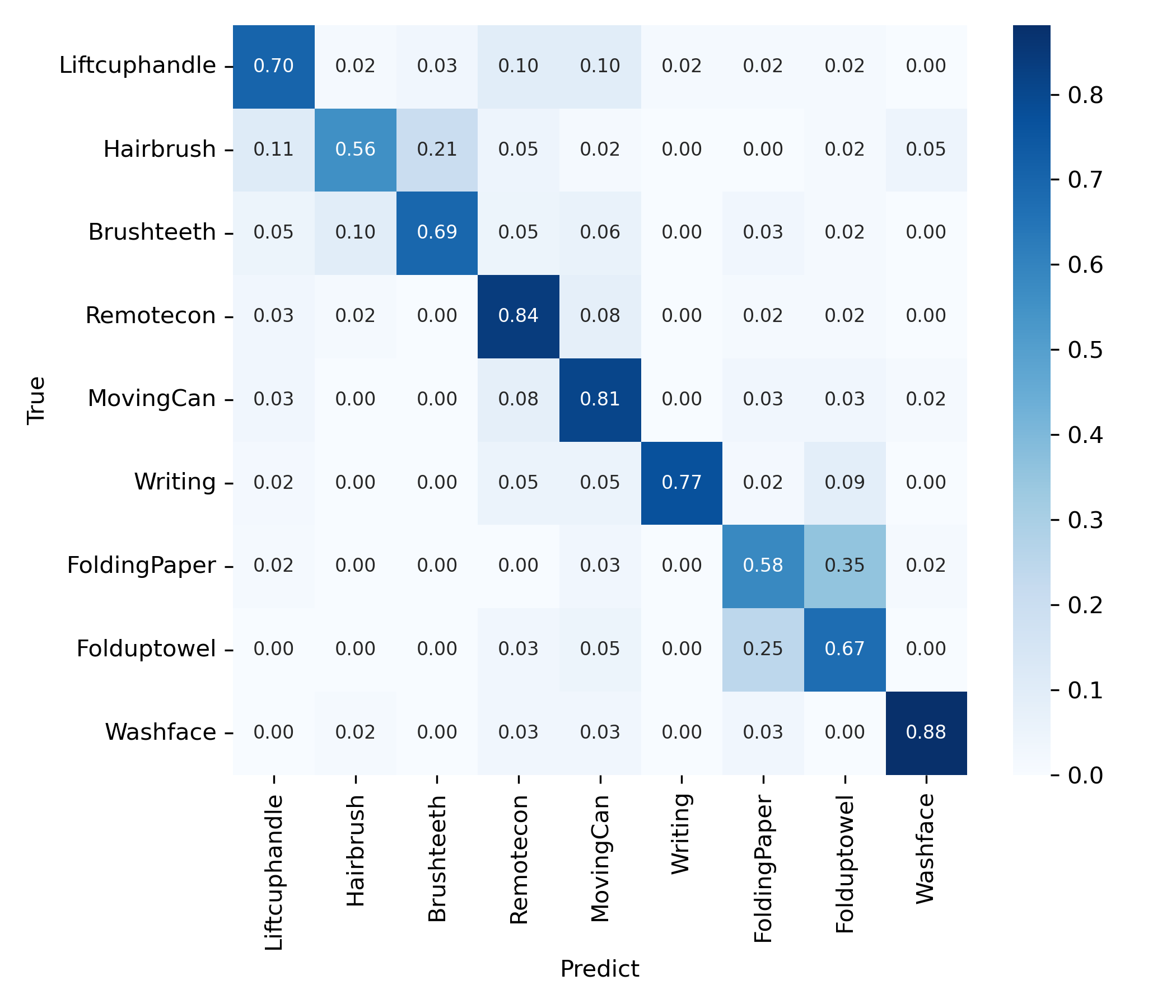}}
\caption{Confusion matrix of Result} 
\label{fig5} 
\end{figure}

Using the training data described in Figure \ref{fig5}, the model was trained and subsequently evaluated on the test data, achieving an accuracy of 72.13\% and a loss of 0.8064.

\subsection{F1 score of segments}

In order to confirm the similarity of segments by class according to the participant group (ND or Stroke), the F1 score was calculated as follows:

\begin{gather}
\text{Precision} = \frac{\text{TP}}{\text{TP} + \text{FP}} \\
\text{Recall} = \frac{\text{TP}}{\text{TP} + \text{FN}} \\
F_1 = 2 \cdot \frac{\text{Precision} \cdot \text{Recall}}{\text{Precision} + \text{Recall}} \\
F_1 = 2 \cdot \frac{\tfrac{\text{TP}}{\text{TP}+\text{FP}} \cdot \tfrac{\text{TP}}{\text{TP}+\text{FN}}}
              {\tfrac{\text{TP}}{\text{TP}+\text{FP}} + \tfrac{\text{TP}}{\text{TP}+\text{FN}}} \\
F_1 = \frac{2 \, \text{TP}}{2\,\text{TP} + \text{FP} + \text{FN}}
\end{gather}

\begin{figure}[h]
\centerline{\includegraphics[width=\columnwidth]{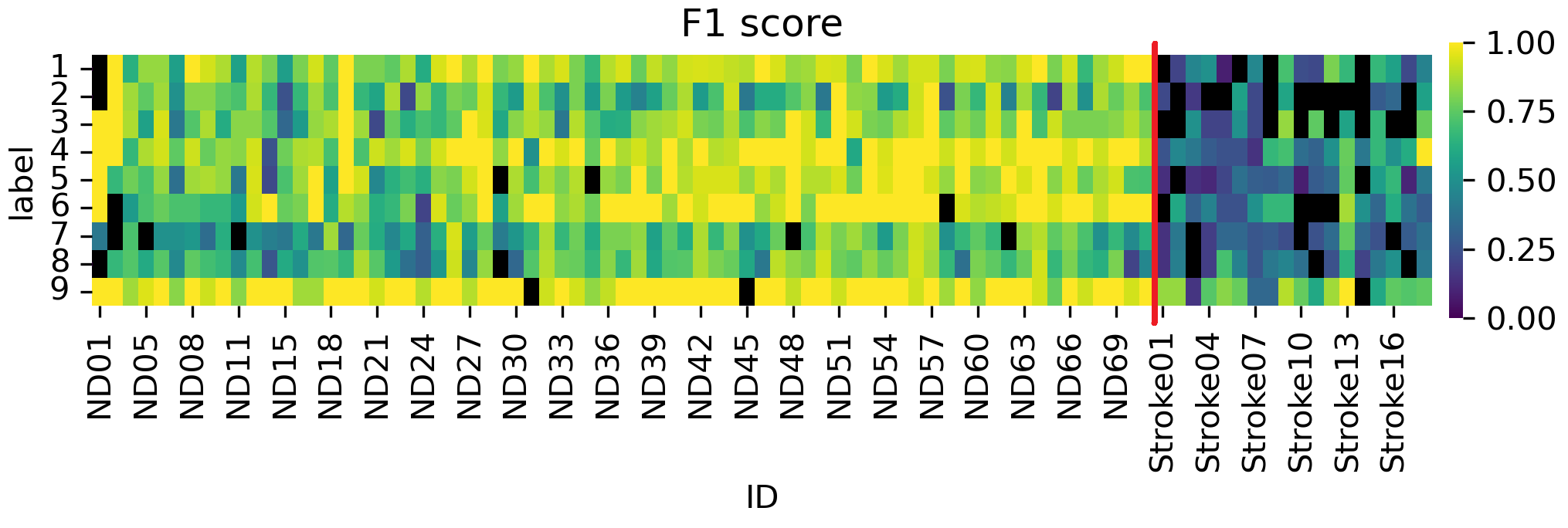}}
\caption{F1 score per every participant and label. The black region means NaN value.}
\label{fig6}
\end{figure}

In the Figure \ref{fig6}, there are F1 scores for each label across all the participants, the regions marked in black correspond to NaN values. The average of F1 score of ND data is 0.8112, while that of Stroke data is 0.4488. It can be observed that the F1 scores for the ND data are generally higher than those for the Stroke data, so that the similarity within the same class is higher in ND group than Stroke group.

\subsection{Ablation Study}
\subsubsection{Model}

The model achieved approximately 82.9\% accuracy in recognizing actions performed by non-disabled participants, indicating that the model is functioning properly. For stroke action recognition, we observed that training the model using both Stroke and ND data yields better accuracy compared to training with Stroke data alone. Therefore, incorporating ND data during training improves the recognition performance over using only Stroke data.

\begin{table}[h]
\centering 
\caption{Ablation study of the type of train data}
\resizebox{\columnwidth}{!}{
\begin{tabular}{c!{\vrule width 1pt}rr|rr}
\Xhline{2\arrayrulewidth}
\multicolumn{1}{c!{\vrule width 1pt}}{\multirow{2}{*}{Tested Dataset}} & \multicolumn{2}{c|}{Accuracy (\%)} & \multicolumn{2}{c}{Loss} \\ 
\multicolumn{1}{c!{\vrule width 1pt}}{} & \multicolumn{1}{c}{IMU} & \multicolumn{1}{c|}{Skeleton} & \multicolumn{1}{c}{IMU} & \multicolumn{1}{c}{Skeleton} \\ 
\Xhline{2\arrayrulewidth}
ND & 82.9268 & 27.0509 & 0.02435 & 16.44946 \\ 
Stroke & 39.7959 & 22.4489 & 1.25016 & 7.08314 \\ 
ND+Stroke & 73.9526 & 26.2295 & 0.03237 & 12.26086 \\ 
\Xhline{2\arrayrulewidth}
\end{tabular}
}
\label{tab6} 
\end{table}

\subsubsection{Multimodal}

The performance of the multimodal HAR was lower than that of the unimodal HAR trained solely on IMU data. However, inspection of the confusion matrix for the skeleton-based unimodal model revealed that it did not assign a single label to all segments. Instead, it produced varying labels across segments, indicating that the model had learned the inherent patterns of the data.

\begin{figure}[h]
    \centerline{\includegraphics[width=0.8\columnwidth]{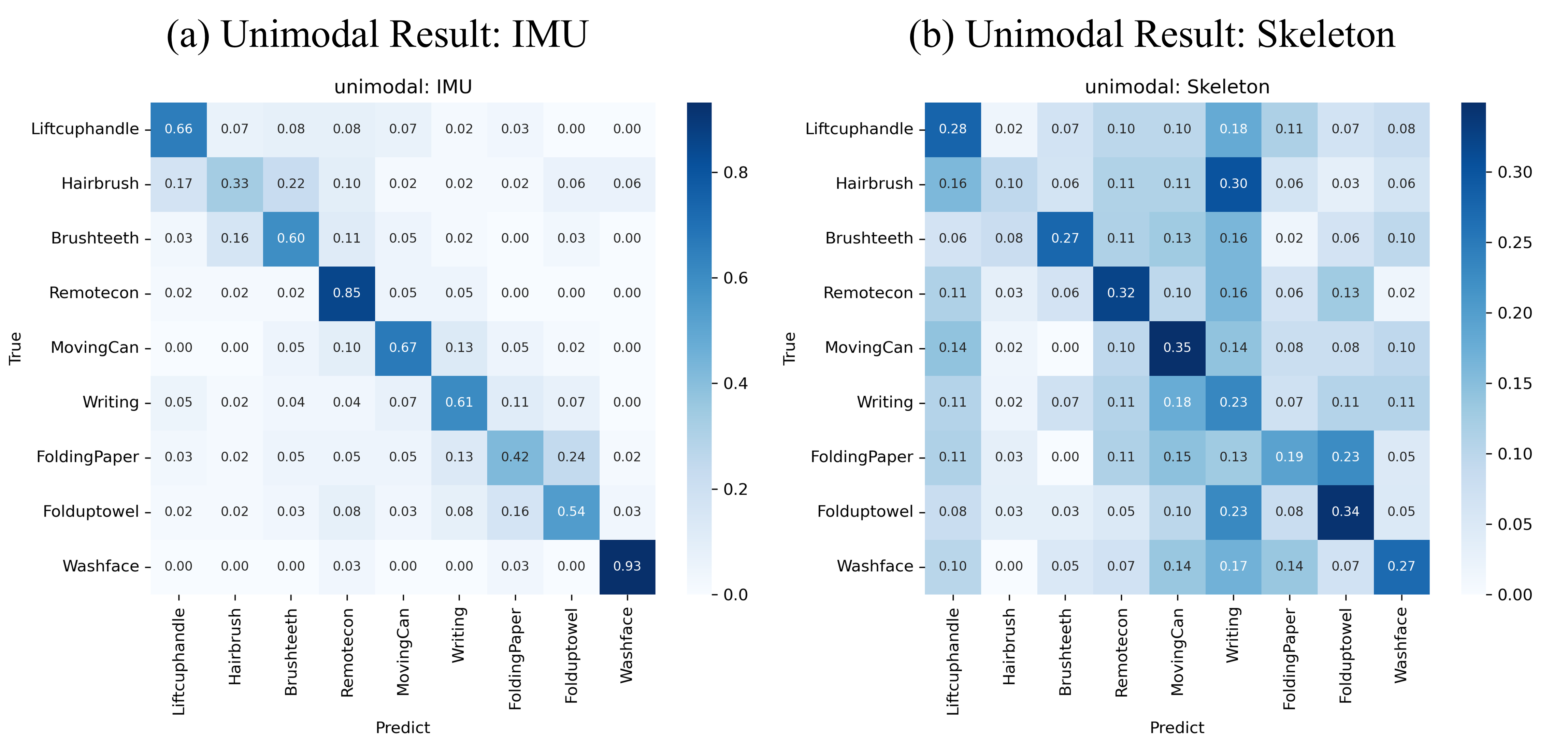}}
    \caption{Confusion matrix of Ablation studies. (a) Confusion matrix of IMU unimodal trained model, (b) Confusion matrix of Skeleton unimodal trained model.}
    \label{fig7}
\end{figure}

\subsubsection{Converting Time Series Data as Image}
To confirm whether it is appropriate to process time series data as a 3-channel 2D image, the accumulation was compared with the ViT model learned by processing sensor data like a 1-channel grayscale 2D image.

\begin{figure*}[!t] 
\centerline{\includegraphics[width=0.9\textwidth]{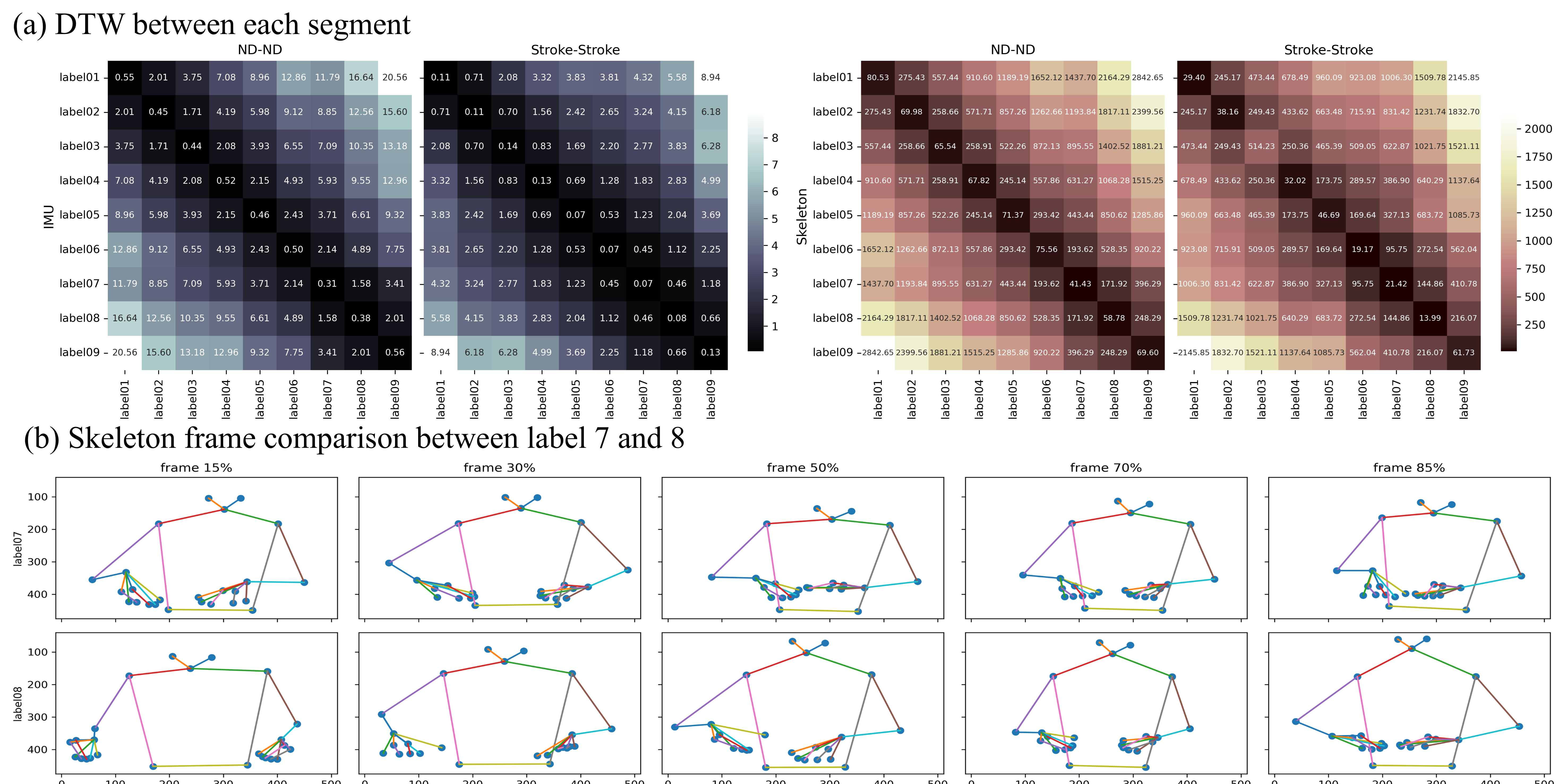}}
\caption{DTW results and comparing frames. (a) DTW values as a heatmap for ND-ND IMU, Stroke-Stroke IMU, ND-ND Skeleton, Stroke-Stroke Skeleton, from left to right. (b) Comparision of skeleton frames at 15\%, 30\%, 50\%, 70\%, and 85\% of each segment's length for labels 7 and 8. } 
\label{fig8} 
\end{figure*}

\begin{table}[h]
\centering
\caption{Performance comparison between raw and modified IMU data.}
\begin{tabular}{l|cc}
\Xhline{2\arrayrulewidth}
 & IMU-raw (1D) & IMU-modified (3D) \\
\Xhline{2\arrayrulewidth}
Accuracy (\%) & 75.41 & 76.26 \\
Loss & 0.7036 & 0.6983 \\
\Xhline{2\arrayrulewidth}
\end{tabular}
\label{tab7}
\end{table}

As shown in Table \ref{tab7}, it can be confirmed that training with IMU data modified into a 3-channel 2D image is also effective.

\subsection{DTW for all segment}

To calculate the Dynamic Time Warping (DTW) between two segments of multivariate time series data, we computed the DTW for each channel along the time axis and then averaged them like Algorithm \ref{algorithm:DTW}.  When examining the DTW of the IMU data, we observed that labels 2, 3, and 4; 3, 4, and 5; and 4, 5, and 6, actions primarily involving the hands, were grouped together, resulting in DTW values of 1 or less, even though only two sensors were used.

\begin{algorithm}[h]
\caption{Multivariate DTW}
\label{algorithm:DTW}
\begin{algorithmic}[1]
\State \textbf{Input:} Multivariate sequences 
\State \quad $A = \{A^{(1)}, \dots, A^{(C)}\}$, $B = \{B^{(1)}, \dots, B^{(C)}\}$
\State \textbf{Initialize:} $\text{total} \gets 0$
\For{$i = 1$ \textbf{to} $C$}
    \For{$j = 1$ \textbf{to} $C$}
        \State Compute DTW distance:
        \State \quad $\operatorname{DTW}(A^{(i)}, B^{(j)}) = 
        \min\left\{
            D(p-1,q),\;
            D(p,q-1),\;
            D(p-1,q-1)
        \right\} + (A^{(i)}_{p} - B^{(j)}_{q})^2$

        \State $\text{total} \gets \text{total} + \operatorname{DTW}(A^{(i)}, B^{(j)})$
    \EndFor
\EndFor
\State \textbf{Output:} $D_{\text{multi}}(A,B) \gets \dfrac{\text{total}}{C}$
\end{algorithmic}
\end{algorithm}

To prevent the model from becoming confused by labels that could be incorrectly perceived as having high similarity, such as these, we included skeleton data, which provides many keypoints. However, as particularly evident in label 3 of the stroke skeleton DTW, some instances of the same action did not have similar similarity values. Since the IMU data can distinguish these well, it confirms the complementary nature of the IMU and skeleton data.

\begin{figure}[h]
\centerline{\includegraphics[width=0.8\columnwidth]{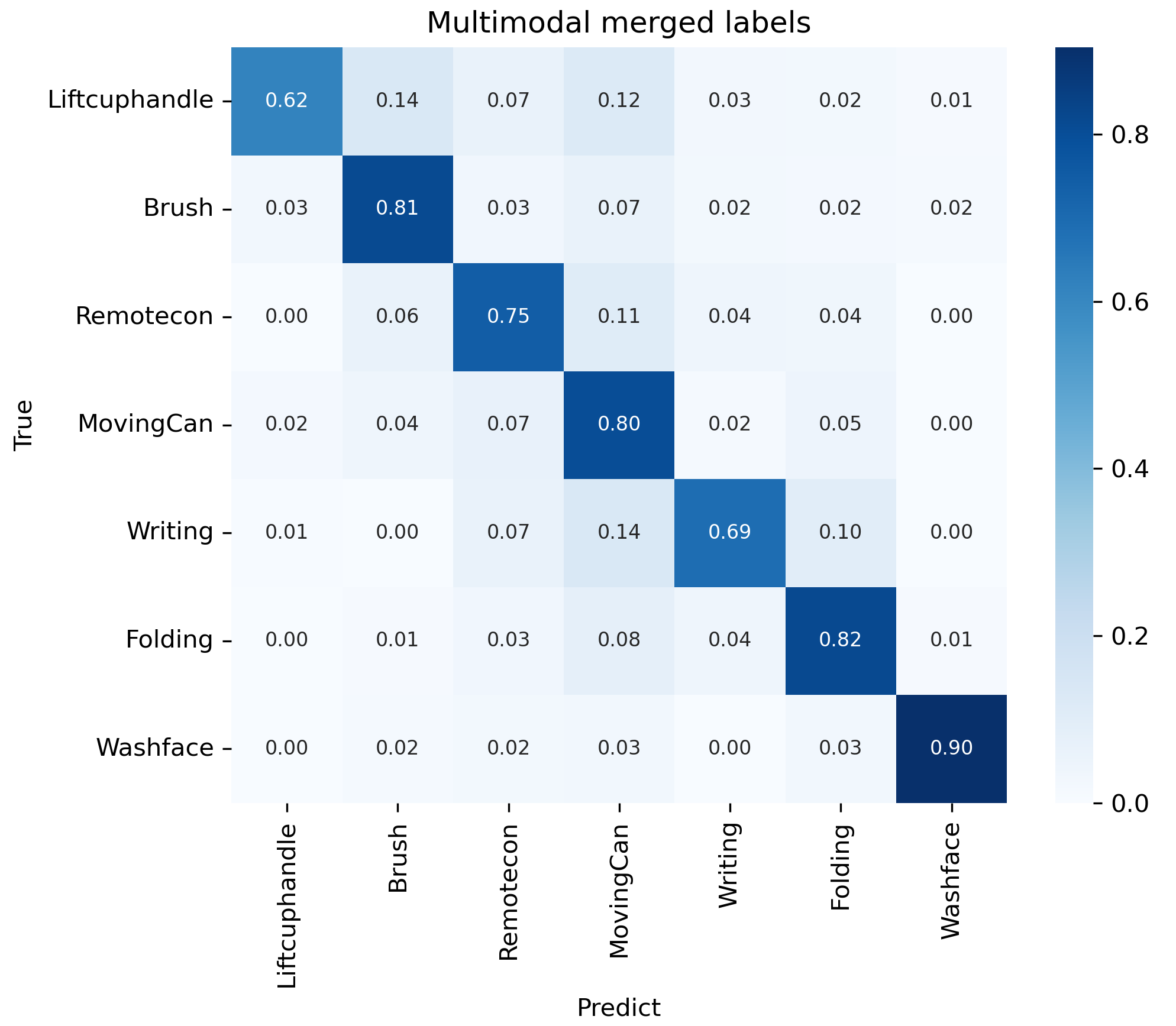}}
\caption{Confusion matrix of merged label result} 
\label{fig9}
\end{figure}

Furthermore, in both ND/Stroke and IMU/Skeleton modalities and classifications, labels 2-3 and 7-8 consistently showed low DTW, indicating high similarity at Figure \ref{fig8}(a). Looking at the skeleton frames at 15\%, 30\%, 50\%, 70\%, and 85\% of each segment's length for labels 7-8 in Figure \ref{fig8}(b), we can confirm that they are similar actions. Since these action labels (brushing and folding) can also be considered similar in real-life daily activities, we treated labels 2-3 and 7-8 as a single action and trained the deep learning model with same hyperparameter. The model trained with 7 labels achieved an accuracy of 78.14\% and a loss of 0.6852, its confusion matrix is shown in Figure \ref{fig9}.

\section{Conclusion}

\label{sec:Conclusion}

\subsection{Limitation}
The characteristic limitations of skeleton data and stroke patient data made action recognition challenging. In this study, we developed a system for collecting indoor upper limb ADL of post-stroke patients and validated the usability of a multimodal motion acquisition and HAR system using data obtained from real experiments. Although multimodal approaches were employed to complement the limitations of unimodal methods, the excessive number of skeleton keypoints and the high similarity among upper limb motions \cite{hardPostStroke} caused finger data to be perceived as similar movements by the model. Consequently, the performance of multimodal HAR was lower than that of IMU-based unimodal HAR. However, analysis of the confusion matrix of the skeleton unimodal model revealed that the model did not converge to a single label but instead produced diverse labels depending on the segment, indicating that it had learned the underlying trends of the data. In future work, we plan to design model architectures capable of faster and more pertinent classification of skeleton data and pursue their practical implementation.

\subsection{Future Works}
This study showed that human action recognition of post-stroke patients is possible using a multimodal deep learning model. In future studies, this well be developed and expanded to not only simple action recognition but also feedback such as assessment that contributes to domiciliary rehabilitation. So we plan to build a complete domiciliary ADL monitoring system for post-stroke patients. Furthermore, by exploring alternative approaches to skeleton-based learning, we seek to improve model performance and thereby enhance the action classification capability of multimodal HAR. In addition, we intend to expand beyond ADL tasks by incorporating ROM exercises, thereby demonstrating the broader applicability of the proposed approach not only to ADL but also to ROM and, ultimately, to rehabilitation action recognition in general.

\section*{Acknowledgment}
This study was supported by the Translational Research Center for Rehabilitation Robots (\#NRCTR-EX23008), National Rehabilitation Center, Ministry of Health and Welfare, Korea and the Gachon University research fund of 2024(GCU-202400480001).

\ifCLASSOPTIONcaptionsoff
  \newpage
\fi

\bibliographystyle{IEEEtran}
\bibliography{IEEEabrv,Bibliography}

\vfill

\end{document}